\documentclass[10pt,twocolumn,letterpaper]{article}

\usepackage[pagenumbers]{cvpr} 

\usepackage{mathtools}
\usepackage{amsmath}
\usepackage{array,multirow,graphicx}
\usepackage{url}            
\usepackage{booktabs}       
\usepackage{amsfonts}       
\usepackage{nicefrac}       
\usepackage{microtype}      
\usepackage{bm, eucal}
\usepackage[table]{xcolor}   
\usepackage{pifont}
\newcommand{\cmark}{\ding{51}}%
\newcommand{\xmark}{\ding{55}}%
\definecolor{darkergreen}{RGB}{21, 152, 56}
\usepackage{multicol}
\usepackage{wrapfig}
\usepackage{caption}
\usepackage{tikz}
\newcommand*{\circled}[1]{\lower.7ex\hbox{\tikz\draw (0pt, 0pt)%
    circle (.5em) node {\makebox[1em][c]{\small #1}};}}

\definecolor{cvprblue}{rgb}{0.21,0.49,0.74}
\usepackage[pagebackref,breaklinks,colorlinks,citecolor=cvprblue]{hyperref}

\title{REACT: Recognize Every Action Everywhere All At Once}

\author{Naga VS Raviteja Chappa$^{1}$, Pha Nguyen$^{1}$, Page Daniel Dobbs$^{1}$,Khoa Luu$^{1}$\\
$^{1}$University of Arkansas\\
\tt\small \{nchappa, panguyen, pdobbs, khoaluu\}@uark.edu,  
}

\begin{document}
\maketitle

\begin{abstract}

Group Activity Recognition (GAR) is a fundamental problem in computer vision, with diverse applications in sports video analysis, video surveillance, and social scene understanding. Unlike conventional action recognition, GAR aims to classify the actions of a group of individuals as a whole, requiring a deep understanding of their interactions and spatiotemporal relationships. To address the challenges in GAR, we present REACT (\textbf{R}ecognize \textbf{E}very \textbf{Act}ion Everywhere All At Once), a novel architecture inspired by the transformer encoder-decoder model explicitly designed to model complex contextual relationships within videos, including multi-modality and spatio-temporal features.
Our architecture features a cutting-edge Vision-Language Encoder block for integrated temporal, spatial, and multi-modal interaction modeling. This component efficiently encodes spatiotemporal interactions, even with sparsely sampled frames, and recovers essential local information. Our Action Decoder Block refines the joint understanding of text and video data, allowing us to precisely retrieve bounding boxes, enhancing the link between semantics and visual reality. At the core, our Actor Fusion Block orchestrates a fusion of actor-specific data and textual features, striking a balance between specificity and context.
Our method outperforms state-of-the-art GAR approaches in extensive experiments, demonstrating superior accuracy in recognizing and understanding group activities. Our architecture's potential extends to diverse real-world applications, offering empirical evidence of its performance gains. This work significantly advances the field of group activity recognition, providing a robust framework for nuanced scene comprehension.

\end{abstract}    

\section{Introduction}\label{sec:intro}

Group activity recognition (GAR) has emerged as a pivotal challenge in the realm of computer vision, with diverse applications spanning sports video analysis, surveillance systems, and social scene understanding~\cite{wang2016temporal, carreira2017quo, wang2018non, Ranasinghe_2022_CVPR}. Unlike conventional action recognition techniques that focus on individual actions, GAR aims to identify and classify the collective actions of a group of individuals within a video segment, necessitating a deeper understanding of the intricate interactions between these actors~\cite{wang2016temporal, carreira2017quo, wang2018non, Ranasinghe_2022_CVPR}. This complexity poses significant challenges that hinder the development of practical GAR solutions.

\begin{figure}[t!]
    \centering
    \includegraphics[width=0.47\textwidth]{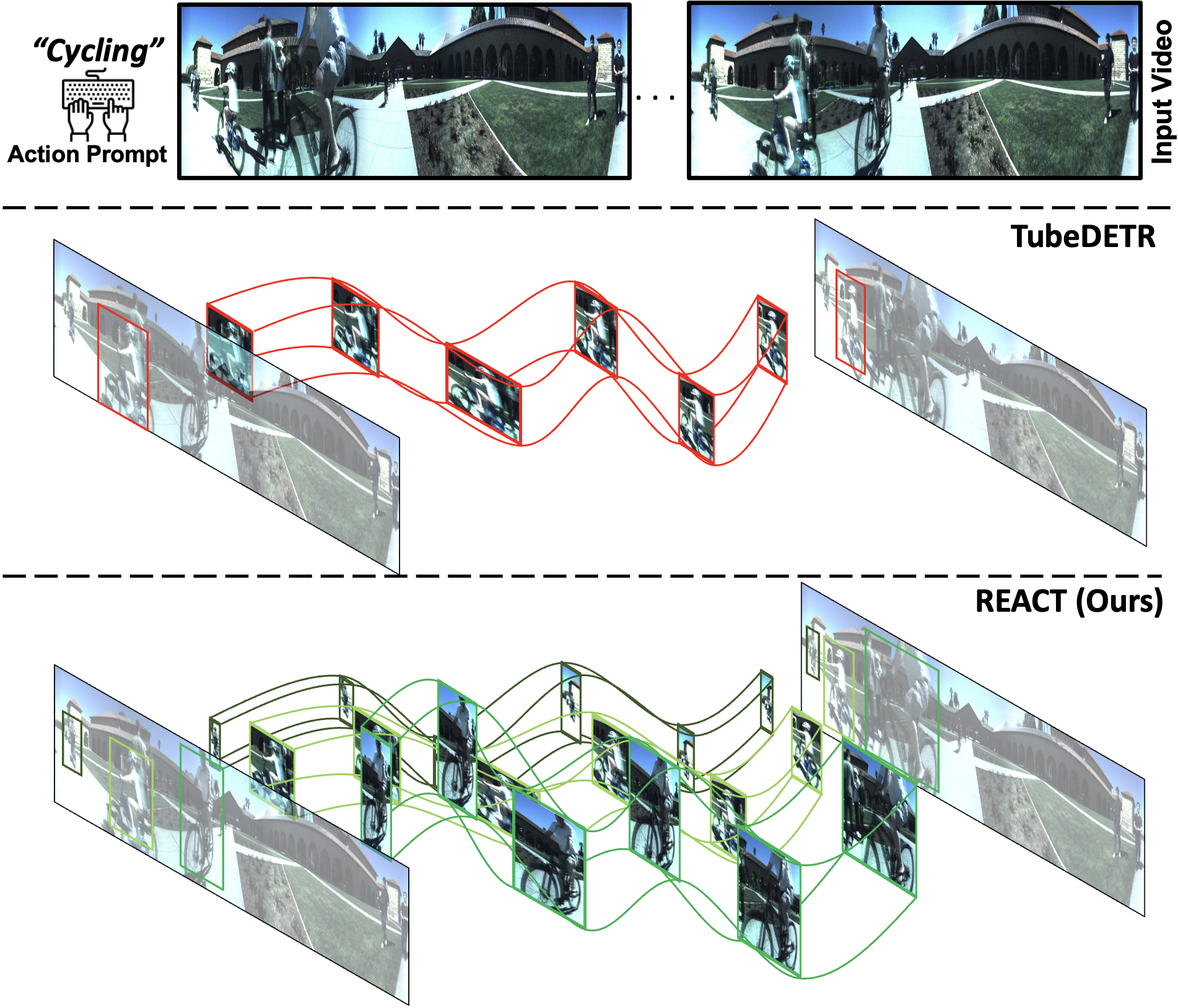}
    \caption{
    {An example of the response of \textit{REACT model} to the user's \textit{input query}. The user provides a video sequence and an action prompt. Then, the model outputs all the requested actions in the scene by localizing the corresponding actors and provides the overall group activity. \textbf{Best viewed in color and zoom in.}}}
    \label{fig:page1}
\end{figure}

Current GAR methods typically rely on ground-truth bounding box annotations and action class labels for training and evaluation~\cite{ibrahim2016hierarchical, wu2019learning, hu2020progressive, gavrilyuk2020actor, pramono2020empowering, ehsanpour2020joint, yan2020higcin, yuan2021learning, li2021groupformer}. These methods extract actor features and their spatio-temporal relationships from bounding box annotations, and then aggregate these features to form a group-level video representation for classification~\cite{ibrahim2016hierarchical, wu2019learning, hu2020progressive, gavrilyuk2020actor, pramono2020empowering, ehsanpour2020joint, yan2020higcin, yuan2021learning, li2021groupformer}. However, this reliance on annotations significantly restricts the scalability and applicability of GAR methods.
To overcome these limitations, some approaches combine person detection and GAR tasks, simultaneously training on bounding box annotations~\cite{bagautdinov2017social, zhang2019fast}. Another approach, termed weakly supervised (WSGAR) learning, aims to eliminate the need for actor-level annotations~\cite{yan2020social, kim2022detector}.

\begin{figure}[t!]
    \centering
    \includegraphics[width=0.475\textwidth]{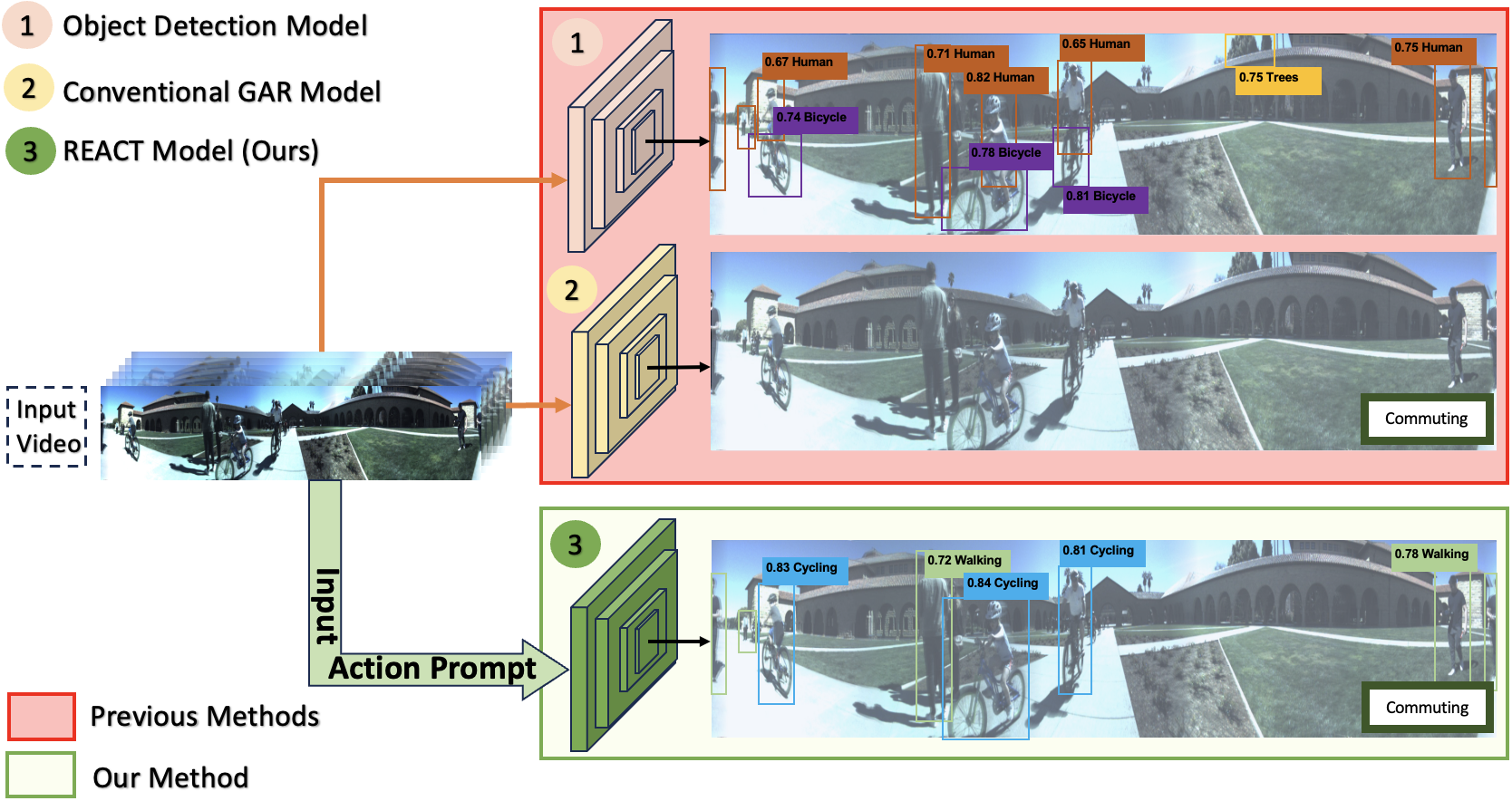}
    
    \caption{
    {Comparison between prior methods and our approach. Prior methods do a single classification/detection task while fully supervised, whereas our approach performs group activity classification and query-based action detection simultaneously. \textbf{Best viewed in color and zoom in.} }}
    \label{fig:motivation}
 \end{figure}

Yan~\etal~\cite{yan2020social} proposed the WSGAR learning approach, which utilizes a pre-trained detector to generate actor box suggestions and learns to disregard irrelevant possibilities. However, this method faces challenges when actors are occluded or poorly detected. Kim~\etal~\cite{kim2022detector} introduced a detector-free method that captures actor information using partial contexts of token embeddings, but this method is restricted to scenarios with movement transitions between frames. Additionally, neither method considers the consistency of temporal information across different tokens.

\subsection{Our Contributions in this Work}
\begin{figure*}[t!]
    \centering
    \includegraphics[width=1.0\textwidth]{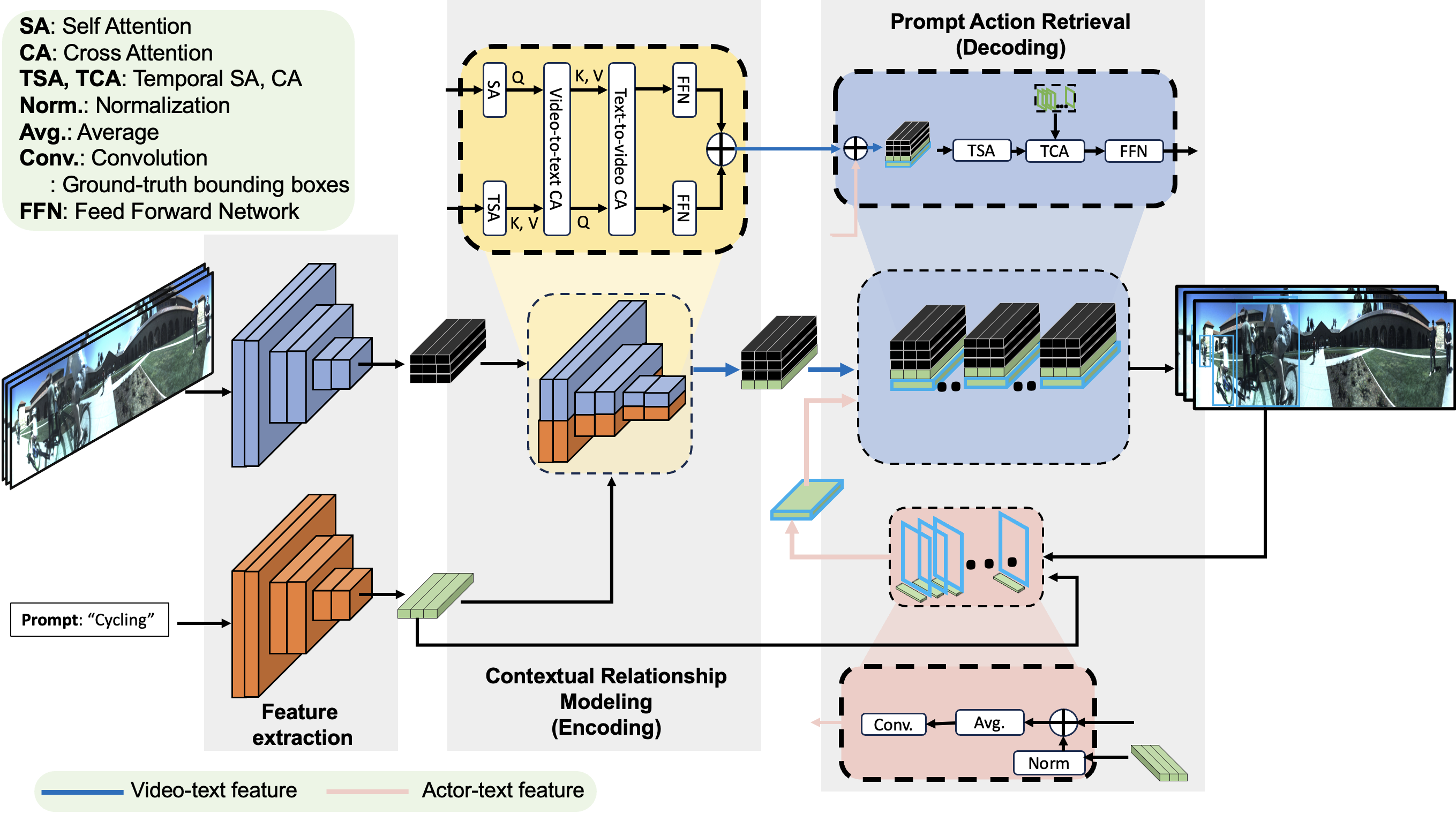}
    \put(-476, 114){\footnotesize \rotatebox{30}{Input Video}}
    \put(-467, 108){\footnotesize \rotatebox{30}{(Sampled)}}
    \put(-420,190){\footnotesize\textbf{Visual}}
    \put(-396,190){\footnotesize\textbf{backbone}}
    \put(-416,111){\footnotesize\textbf{Text}}
    \put(-399,111){\footnotesize\textbf{backbone}}
    \put(-345,86){\footnotesize$\bm{t_f}$}
    \put(-345,172){\footnotesize$\bm{v_f}$}
    \put(-128, 241){\footnotesize\textbf{$\bm{b_g}$}}
    \put(-321, 271){\footnotesize\textbf{Vision-Language}}
    \put(-307,264){\footnotesize\textbf{Encoder}}
    \put(-187, 130){\small\textbf{Action Decoder}}
    \put(-167, 122){\small\textbf{Block}}
    \put(-204, 40){\small\textbf{Actor Fusion Block}}
    \put(-230,174){\footnotesize$\bm{\widehat{vt_f}}$}
    \put(-237,115){\footnotesize$\bm{\widehat{bt_f}}$}
    \put(-65,185){\footnotesize$\bm{\{\widehat{b_i}\}_{i=1,..,n}}$}
    \put(-489,213){\small\textbf{$\bm{b_g}$}}
    \caption{
    \textbf{Overall architecture of the proposed REACT network.} The visual and textual representation learning components from our approach incorporate multi-level feature representations. The extracted features are passed through the contextual relationship modeling block to obtain the concatenated multi-modality features. Then, it is passed through the prompt action retrieval block to obtain the detected bounding boxes based on the prompt.} \label{fig:overall}
\end{figure*}
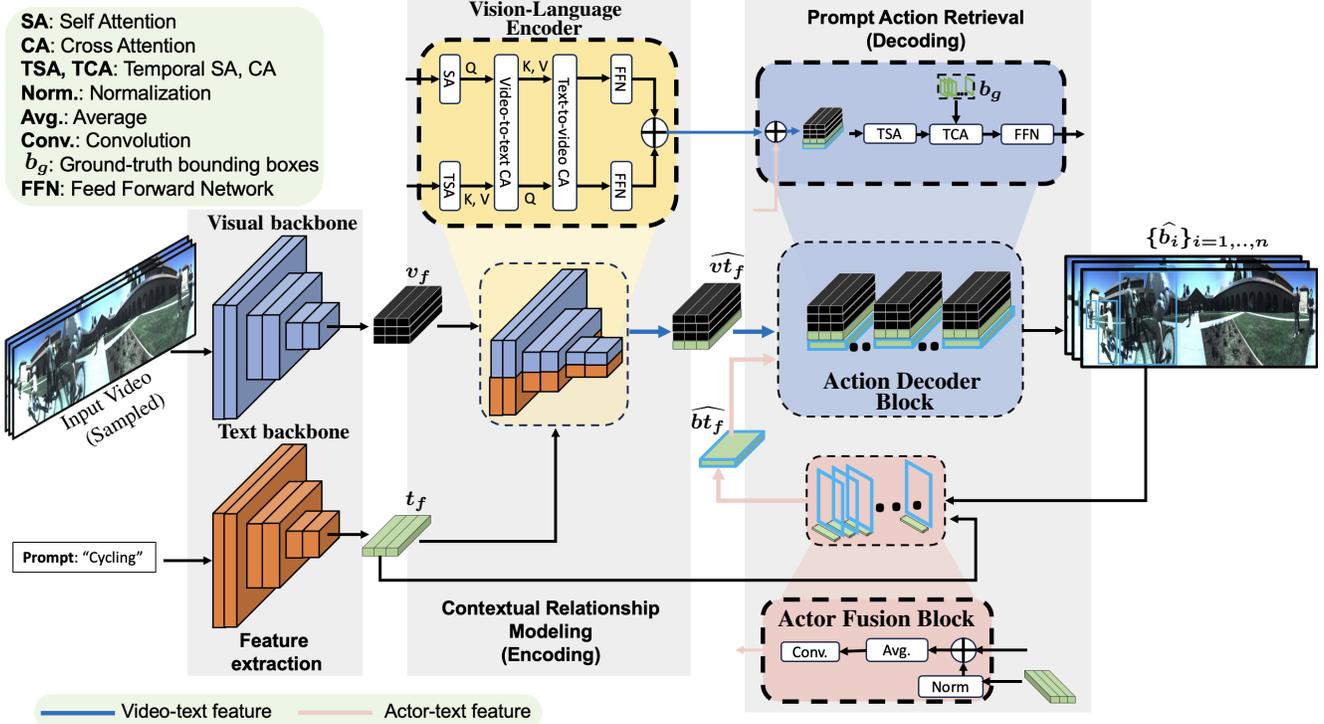

Inspired by attention mechanisms in video contexts, we introduce a transformer-based architecture, as depicted in~\cref{fig:overall}, to effectively model complex interactions within videos, incorporating temporal, spatial, and multi-modal features. The \textbf{Vision-Language (VL) Encoder block} (\cref{sec:vlencoder}) stands as a pivotal element, proficient in encoding sparse spatial and multi-modal interactions, with a dedicated fast branch for recovering temporal details.
Our architecture integrates a \textbf{Action Decoder Block} (\cref{sec:decoder}) that refines the understanding of text and video data. It balances context and actor-specific details by merging Encoder and Actor Fusion Block outputs. Utilizing temporal spatial attention for discerning temporal and spatial dependencies and temporal cross attention, aligned with ground-truth bounding boxes, ensures precise retrieval of bounding boxes, bridging semantic and visual contexts.
The \textbf{Actor Fusion Block} (\cref{sec:actorfusion}), positioned at the core, orchestrates a harmonious blend of actor-specific information and textual features. Through concatenation and averaging operations, it achieves a holistic representation enriched by convolution operations extracting local patterns. This refined representation contributes to contextually relevant output generation in the Decoder Block.
In practical terms, our model's efficacy is demonstrated in Fig.~\ref{fig:page1} through an example response to a user's action query. The results showcase superior performance in accurately recognizing and understanding group activities compared to state-of-the-art methods. Our experiments provide empirical evidence of the significant performance gains achieved by our framework, affirming its potential for diverse real-world applications.

\section{Related Work}\label{sec:related}

\paragraph{Group Activity Recognition (GAR).}

Group activity recognition (GAR) has emerged as a prominent area of research in computer vision, owing to its diverse applications in video surveillance, human-robot interaction, and sports analysis. GAR aims to identify and classify the collective actions of a group of individuals within a video segment, necessitating a comprehensive understanding of the intricate interactions between these actors. Initial GAR methods relied on probabilistic graphical methods and AND-OR grammar techniques to process extracted features~\cite{amer2012cost,amer2013monte,amer2014hirf,amer2015sum,lan2011discriminative,lan2012social,shu2015joint,wang2013bilinear}. These methods, while effective, faced limitations in capturing complex interactions and temporal dependencies within group activities. With the advent of deep learning, GAR methods underwent a significant transformation. Convolutional neural networks (CNNs) and recurrent neural networks (RNNs) proved particularly adept at learning high-level information and temporal context, leading to substantial improvements in GAR performance~\cite{bagautdinov2017social,deng2016structure,ibrahim2016hierarchical,ibrahim2018hierarchical,li2017sbgar,qi2018stagnet,shu2019hierarchical,wang2017recurrent,yan2018participation}.

Current GAR methods predominantly utilize attention-based models that require explicit character representations to model spatial-temporal relations in group activities~\cite{ehsanpour2020joint,gavrilyuk2020actor,hu2020progressive,li2021groupformer,pramono2020empowering,wu2019learning,yan2020social,yuan2021spatio}. These models employ techniques such as graph convolution networks, spatial and temporal relation graphs, clustered attention, MAC-Loss~\cite{han2022dual} and transformer encoder-based architectures to capture contextual spatial-temporal information and infer actor links. Tamura~\etal~\cite{tamura2022hunting} introduced a framework for recognizing social group activities and identifying group members without relying on heuristic features. This framework embeds the activity and group member information into the features, facilitating efficient identification.


\paragraph{Weakly Supervised Group Activity Recognition (WSGAR).} Traditional GAR approaches rely heavily on ground-truth bounding box annotations and action class labels for training and evaluation, which can be time-consuming and laborious to obtain. Weakly supervised group activity recognition (WSGAR) addresses this challenge by utilizing less restrictive data sources such as activity labels, bounding boxes, or video-level labels~\cite{zhang2021multi,kim2022detector}.

\paragraph{Transformers in Vision.}
Transformers have revolutionized the field of natural language processing (NLP), and their impact has extended to computer vision as well. Vaswani~\etal~\cite{vaswani2017attention} introduced the transformer architecture for sequence-to-sequence machine translation, and since then, it has been successfully applied to various vision tasks. In the video domain, transformers~\cite{han2020mining, arnab2021vivit, li2022uniformer, bertasius2021space,fan2021multiscale,patrick2021keeping,yang2022tubedetr} have demonstrated their effectiveness in learning video representations by utilizing spatial and temporal self-attention. Bertasius~\etal~\cite{bertasius2021space} explored different mechanisms of space and time attention to learn spatiotemporal features efficiently, while Fan~\etal~\cite{fan2021multiscale} employed multiscale feature aggregation to improve the learning performance of features. Patrick~\etal~\cite{patrick2021keeping} introduced a self-attention block that focuses on the trajectory, which tracks the patches of space and time in a video transformer.

\def\GS{{\rm GS}}
\def\LT{{\rm LT}}
\def\l2v{{l}}
\def\v2l{{v}}

\section{The Proposed Method}

As shown in~\cref{fig:overall}, given the input video $\mathbf{X}$ and the textual input $\mathbf{t}$, the REACT model aims to localize all the positions $\mathbf{b}$ in the video $\mathbf{X}$ that consists of the group actions of interest based on the input prompt $\mathbf{t}$. Formally, the REACT model can be formulated as follows:
\begin{align}
    \hat{\mathbf{b}} = \textit{AD}(\mathbb{F}(\textit{VE}(\mathbf{X}),  \textit{TE}(\mathbf{t}))) 
\end{align}
where $\hat{\mathbf{b}} \in [0,1]^{T\times4}$ is the normalized predicted bounding box coordinates, \textit{VE} and \textit{TE} are the visual and textual encoders to extract the feature representations of the video input $\mathbf{X}$, i.e., $\bm{v_f}$ = \textit{VE}($\mathbf{X}$), and the textual prompt $\mathbf{t}$, i.e., $\bm{t_f}$ = \textit{TE}($\mathbf{t}$), and $\mathbb{F}$ is the correlation model to exploit the contextual relationship between visual features $\bm{v_f}$ and the textual features $\bm{t_f}$. Then, the Action decoder(\textit{AD}) will decode the contextual features produced by the correlation model $\mathbb{F}$ to localize all the group action regions of interest. Fig.~\ref{fig:overall} illustrates the proposed framework of REACT.

Contextual Relationship Modeling in GAR involves modeling the multi-modality and spatio-temporal features. Thus, to efficiently cover these correlations, we model the correlation function $\mathbb{F}$ via the attention mechanism. In particular, the correlation model $\mathbb{F}$ is designed as a Transformer network. Our model $\mathbb{F}$ is at the forefront of our proposed approach which comprises vital components for the integrated modeling of temporal, spatial, and multi-modal interactions. It adeptly encodes spatial and multi-modal interactions, efficiently processing sparsely sampled frames. To increase the multi-modal interaction modeling in our Grounded GAR model, we adopt the cross-attention mechanism to model video-to-text and text-to-video correlation. While the video-to-text correlation enables the model to associate visual information within the video frames with corresponding textual descriptions, effectively linking the two modalities, the text-to-video correlation enhances its ability to relate text to specific visual elements or actions within the video. Formally, $\mathbb{F}$ can be formed as follows:
%
  \begin{align}    
  \bm{\widehat{vt_f}} & = \mathbb{F}(\bm{v_f},\bm{t_f})
  = \textit{FFN}(\textit{T2V}(\bm{t_f}, \bm{v_f})( (\textit{V2T}(\bm{v_f}, \bm{t_f}))))
\end{align}
%
where $\bm{\widehat{vt_f}}$ is the shared representation of video and text, \textit{FFN} is a feed-forward neural network, \textit{T2V} and \textit{V2T} are the cross-attention models for the visual and textual features. Thanks to the attention mechanism, it has empowered our correlation model $\mathbb{F}$ to produce a shared representation that captures the contextual relationships between text and video in both spatial and temporal dimensions and encapsulates a joint understanding of both text and video content, enabling the model to bridge the semantic gap between the two modalities effectively.

From the shared contextual feature representations produced by the correlation model $\mathbb{F}$, the action decoder aims to localize all the group action positions of interest $\mathbf{b}$. In practice, the group action positions $\mathbf{b}$ could consist of a list of positions of interest $\mathbf{b} = \{\mathbf{b_1}, \mathbf{b_2}, ..., \mathbf{b_N}\}$ where $N$ is the number of group action positions. For example, in a military mission, the group action of enemies could include the positions of many army groups and adversarial armed devices. Additionally, predicting a particular position of group action requires a broader understanding of the contextual feature representation and the surrounding predicted group action positions. Therefore, to efficiently model the group action decoder, in addition the the shared feature representation, the positions of the current predicted group actions are also taken into account. Formally, the Group Action Decoder can be modeled as follows:
\begin{align}
    \mathbf{\widehat{b_i}} = \textit{AD}(\bm{\widehat{vt_f}}, \textit{AF}(\bm{t_f}, \{\mathbf{b_1}, \mathbf{b_2}, ..., \mathbf{b_N}\}))
    = \textit{AD}(\bm{\widehat{vt_f}}, \bm{\widehat{bt_f}})
\end{align}
%
where \textit{AF} (ActorFusion) is the model that exploits the correlation between the current predicted positions
of group actions and the textual features. Intuitively, the list of group action positions $\{\mathbf{b_1}, .., \mathbf{b_{i-1}}\}$ provide the spatial context of subjects in the video. Meanwhile, the textual features $\bm{t_f}$ represent linguistic aspects of context and semantics related to the group actions and offer a
complementary view to the visual information. The \textit{AF} model distinguishes between capturing detailed spatial information from group action positions and emphasizing the broader action-related information from the textual features. Finally, the ActionDecoder (\textit{AD}) model will predict the position of group action $\mathbf{b_i}$ based on the contextual feature representation $\bm{\widehat{vt_f}}$ and the action fusion feature $\bm{\widehat{bt_f}}$ produced by the \textit{AF} model. In essence, the \textit{AD} model uses a wealth of information from the correlation model $\mathbb{F}$ and fusion features from the \textit{AF} model to associate textual queries $\bm{t_f}$ with specific regions or subjects within the video frames. To efficiently exploit the correlation of these features in \textit{AD}, we design \textit{AD} via the attention mechanism. It enables the \textit{AD} model to precisely identify and highlight the regions within the video that correspond to the described actions.

\subsection{Vision-Language (VL) Encoder}\label{sec:vlencoder}




The VL Encoder typically receives input features from their respective backbone networks in this method. These backbones are specialized models designed to extract meaningful information from set of \textit{L} text data $\bm{t_f} \in \mathbb{R}^{L \times d}$ and video data $\bm{v_f} \in \mathbb{R}^{T \times HW \times d}$ for all \textit{T} frames and \textit{HW} spatial resolution, independently optimizing their representations for their specific modalities.

The VL Encoder receives the features $\bm{t_f}$ and $\bm{v_f}$. It also uses visual features derived from the video, including frames, motion patterns, bounding boxes, and object presence.

The VL Encoder utilizes attention mechanisms to comprehend these features effectively. It begins with self-attention on text, allowing it to weigh the importance of words and understand the context within the text. Simultaneously, it performs temporal self-attention on video features to capture temporal dependencies and dynamics.

The model then progresses to cross-attention, connecting the video and text modalities. The "video-to-text" step focuses on video features and aligns them with relevant text. The "text-to-video" step attends to video elements based on the textual context. This bidirectional interaction ensures a holistic understanding of both modalities.

Finally, the information from self-attention and cross-attention is combined, creating multi-modal features. These features merge enhanced text and video information, bridging the gap between the two modalities. The VL Encoder facilitates joint text and video analysis, benefiting tasks like video captioning and action recognition. It empowers models to grasp contextual relationships between text and video, enhancing multi-modal understanding.

\subsection{Action Decoder (AD) Block}\label{sec:decoder}


The \textit{AD} Block is a crucial part of the architecture responsible for producing meaningful bounding box outputs. It combines information from the encoder and the actor fusion blocks.

First, it concatenates outputs from these sources. The Encoder Block provides a broad multi-modal understanding of text and video, while the Actor Fusion Block adds actor-specific details like normalized text features and actor bounding boxes. This merging process combines contextual understanding with fine-grained actor-specific information.

Then, the \textit{AD} Block employs temporal spatial attention. This mechanism allows the model to consider temporal (sequential) and spatial (object relationships) aspects in video data. It is particularly valuable for tasks like action recognition and actor localization.

Next, it uses temporal cross-attention, incorporating ground-truth bounding box data. Ground-truth bounding boxes precisely locate objects or actors in video frames. This step refines the model's understanding of object and actor movements over time, aligning textual descriptions with the video content.

Finally, the \textit{AD} Block generates bounding boxes as grounding outputs based on text queries. It leverages the information gathered from previous stages, aligning textual descriptions with specific regions or objects in video frames. This output is vital for tasks like actor localization and action recognition, connecting text with visual content to pinpoint regions corresponding to described actions or objects.

\subsection{Actor Fusion (AF) Block}\label{sec:actorfusion}



The \textit{AF} block begins by combining two vital sources of information, including the bounding box data $\hat{\mathbf{b}}$ and the normalized text features $\mathbf{t_f}$. The bounding box data provides precise location details for objects or actors in video frames, enhancing the model's spatial understanding.

Simultaneously, normalized text features offer textual descriptions and contextual information related to actions or objects in the video, providing a linguistic perspective.

After combining these sources, the \textit{AF} block calculates an average over the concatenated features. This averaging serves multiple purposes. It creates a balanced representation that includes actor-specific details (from the bounding box) and semantic context (from the text features). Additionally, it emphasizes action-related information, aligning with the action recognition task.

The \textit{AF} block uses convolution operations to refine the feature representation further (presented as $\widehat{\mathbf{bt_f}}$). Convolutions are essential in deep learning for capturing local data patterns. They extract and highlight relevant spatial and temporal features from the combined actor-specific information in this context. These convolutional operations are the key to improving the representation before passing it back to the \textit{AD} Block.

\subsection{Training Loss}
The input training data is in the form of a set of frames, where each set of frames belongs to a video, which is annotated with a text label representing the group activity and individual actions $s$ and the corresponding set of bounding boxes $b$.  

We train our architecture with a linear combination of four losses \\[-17pt]
\begin{align}
\label{eq:objective}
\mathcal{L} = \lambda_{\mathcal{L}_1}\mathcal{L}_{\mathcal{L}_1}(\hat{b}, b) + \lambda_{gIoU}\mathcal{L}_{gIoU}(\hat{b}, b)
\end{align}
\noindent where $b \in [0,1]$ denotes the normalized ground truth box coordinates and $\hat{b}$ the predicted bounding boxes.
Finally, different $\lambda_{\bullet}$ are scalar weights of the individual losses. 
$\mathcal{L}_{\mathcal{L}_1}$ is a $\mathcal{L}_1$ loss on bounding box coordinates. 
$\mathcal{L}_{gIoU}$ is a generalized ``intersection over union" (IoU) loss~\cite{rezatofighi2019generalized} on the bounding boxes. 
Both $\mathcal{L}_1$ and $\mathcal{L}_{gIoU}$ are used for spatial and temporal grounding.  
Losses are computed at each layer of the decoder following~\cite{carion2020end}.
\subsection{Inference}
During the testing phase, we employ distinct operations for individual action and social group predictions. For individual actions, a softmax operation is applied to the predictions from cross-entropy functions, and a sigmoid operation is performed on predictions from binary cross-entropy functions. Our approach follows a hierarchical strategy, initiating label selection from the first partition and progressing to subsequent ones in the hierarchy only if the "Other" class is predicted. In the context of social group prediction, we leverage the predicted counts of individual actions and social groups. To deduce the activity of each predicted social group, we adopt a strategy where the social activity label is determined by the most frequent action labels among its members.

\subsection{Discussion}

\noindent\textbf{Need for concatenated attention?} The method described earlier, which we will call "concatenation-based fusion," involves the combination of features. It performs self-attention on features from individual sources and then computes cross-attention across different sources. It is in contrast to "cross-attention-based fusion." With flexibility and fewer assumptions about the data's spatial pattern, the Transformer model offers significant modeling freedom. Compared to cross-attention-based fusion, concatenation-based fusion is more efficient due to operation sharing and reduces model parameters through weight sharing. Weight sharing is a critical design aspect for symmetric metric learning between two data branches. In concatenation-based fusion, we implement this property in feature extraction and feature fusion. In a nutshell, concatenation-based fusion enhances both efficiency and performance.

\noindent\textbf{What if we use a query-based decoder?} Many transformer-based models in vision tasks draw inspiration from the vanilla Transformer decoder. They incorporate a learnable query to extract specific target features from the encoder. For instance, in \cite{carion2020end}, they use object queries; in \cite{yan2021learning}, they refer to target queries. However, our experimental findings reveal that a query-based decoder faces challenges, including slower convergence and subpar performance. The vanilla Transformer decoder, which is fundamentally a generative model, may not be the ideal choice for classification tasks. Furthermore, employing a single universal target query for all types of objects could potentially create a performance bottleneck. It is worth emphasizing that REACT primarily functions as an "encoder" model within the conventional Transformer encoder-decoder framework.

\section{Experiment Results}


\subsection{Experiment Settings}

\noindent\textbf{Upstream task} 
We consistently used the RoBERTa~\cite{liu2019roberta} model throughout all the experiments to extract the textual features, whereas the visual backbone varied for different experiments to have fair comparisons. We use hyper-parameters \textit{T} = 15 (JRDB-PAR) \& 8 (Volleyball), \textit{N}= 6 (JRDB-PAR) \& 8 (Volleyball), $\lambda_{\mathcal{L}_1}$ = 5, $\lambda_{gIoU}$ = 2 and \textit{d}=256. We initialized temporal attention weights randomly, while spatial attention weights were initialized using a ViT model trained self-supervised over ImageNet-1K \cite{imagenet}. This initialization scheme facilitated faster convergence of space-time ViT, as seen in the supervised setting \cite{gberta_2021_ICML}. We trained using an Adam optimizer \cite{kingma15adam} with a learning rate of $5\times10^{-4}$, scaled using a cosine schedule with a linear warm-up over five epochs \cite{Steiner2021HowTT, chen2021mocov3}. Additionally, we applied weight decay scaled from 0.04 to 0.1 during training. 

\noindent\textbf{Downstream tasks} We trained a linear classifier on our pretrained backbone. During training, the backbone was frozen, and we trained the classifier for 100 epochs with a batch size of 32 on a single NVIDIA-V100 GPU using SGD with an initial learning rate of 1e-3 and a cosine decay schedule. We also set the momentum to 0.9. 

\subsection{Dataset Details}
\noindent\textbf{Volleyball Dataset}~\cite{ibrahim2016hierarchical}: With 55 videos and 4,830 labeled clips (3,493 training, 1,337 testing), this dataset includes annotations for individual actions and group activities with bounding boxes. In WSGAR experiments, we focus solely on group activity labels, excluding individual action annotations. Evaluation metrics include Multi-class Classification Accuracy (MCA) and Merged MCA, ensuring fair comparisons with existing methods.

\noindent\textbf{JRDB-PAR Dataset}~\cite{han2022panoramic}: Featuring 27 action, 11 social groups, and seven global activity categories, this dataset comprises 27 videos (20 training, seven testing) with 27,920 frames and 628k human bounding boxes. Uniformly sampled keyframes (every 15 frames) are used for annotation and evaluation. Precision ($\mathcal{P}_g$), recall ($\mathcal{R}_g$), and F1-score ($\mathcal{F}_g$) are the evaluation metrics for social group activity recognition, treated as a multi-label classification problem~\cite{han2022panoramic}.
\subsection{Classification Task Evaluation}
\noindent\textbf{JRDB-PAR dataset.} 
In our evaluation on the JRDB-Par dataset, we conducted a comparative analysis of our proposed method against leading approaches in both Group Activity Recognition (GAR) and Weakly Supervised Group Activity Recognition (WSGAR). The assessment covered fully supervised and weakly supervised scenarios, and the results are summarized in Table~\ref{tab:jrdbpar}.
In the fully supervised setting, our method demonstrated significant superiority over existing social group activity recognition frameworks across all metrics. Meanwhile, in the weakly supervised setting, our approach outperformed current GAR and WSGAR methods by noteworthy margins, achieving \textit{1.2} for $\mathcal{P}_g$, \textit{1.7} for $\mathcal{R}_g$, and \textit{2.3} for $\mathcal{F}_g$. We also explored the impact of different backbones—ResNet-18, ViT-B/16, and ViT-B/32—on dataset performance. ViT-B/16 emerged as the most effective, as detailed in Table~\ref{tab:jrdbpar} and further discussed in the {Appendix}. Despite the commendable performance of these models in WSGAR, our proposed approach consistently outshone them.

\begin{table}[!t]
\caption{Comparison with the state-of-the-art methods on the Volleyball dataset~\cite{ibrahim2016hierarchical}. MCA and MPCA represent Mean Class Accuracy and Mean Per Class Accuracy. The \textit{green} line indicates the best result.}

\begin{center}
\begin{tabular}{>{\arraybackslash}m{2.4cm}| >  {\centering\arraybackslash}m{2.3cm} >{\centering\arraybackslash}m{0.9cm}>{\centering\arraybackslash}m{1.0cm}}

\hline
\textbf{Method}                             & \textbf{Backbone}              & \textbf{MCA}   & \textbf{Merged MCA}\\ [0.3ex]
\hline
\multicolumn{4}{c}{\textbf{Fully supervised}} \\
\hline
SSU~\cite{bagautdinov2017social}                 & Inception-v3          & 89.9  & - \\
PCTDM~\cite{yan2018participation}                & ResNet-18             & 90.3  & 94.3\\ %
StagNet~\cite{qi2018stagnet}                     & VGG-16                & 89.3  & - \\
ARG~\cite{wu2019learning}                        & ResNet-18             & 91.1 & {95.1}\\ %
CRM~\cite{azar2019convolutional}        & I3D                   & 92.1  & - \\
HiGCIN~\cite{yan2020higcin}                      & ResNet-18             & 91.4  & - \\
AT~\cite{gavrilyuk2020actor}                     & ResNet-18             & 90.0 & 94.0\\ %
SACRF~\cite{pramono2020empowering}               & ResNet-18             & 90.7 & 92.7   \\ %
DIN~\cite{yuan2021spatio}                        & ResNet-18             & {93.1}  & {95.6} \\ %
TCE+STBiP~\cite{yuan2021learning}        & VGG-16                & {94.1}  & - \\
GroupFormer~\cite{li2021groupformer}                      & Inception-v3             & {94.1}  & - \\

\hline
\multicolumn{4}{c}{\textbf{Weakly supervised}} \\
\hline 
PCTDM~\cite{yan2018participation}               & ResNet-18             & 80.5  & 90.0\\
ARG~\cite{wu2019learning}                       & ResNet-18             & 87.4  & 92.9\\
AT~\cite{gavrilyuk2020actor}                    & ResNet-18             & 84.3  & 89.6\\
SACRF~\cite{pramono2020empowering}              & ResNet-18             & 83.3  & 86.1  \\
DIN~\cite{yuan2021spatio}                       & ResNet-18             & 86.5  & 93.1\\
SAM~\cite{yan2020social}                        & ResNet-18             & 86.3  & 93.1\\

DFWSGAR~\cite{kim2022detector}                  & ResNet-18     & 90.5  & 94.4\\
SoGAR~\cite{chappa2023sogar}                    & ResNet-18     & 91.8 & 94.5  \\
SPARTAN~\cite{chappa2023group}                  & ViT-B/16      & 92.9 & 95.6 \\
SoGAR~\cite{chappa2023sogar}                    & ViT-B/16     & 93.1 & 95.9  \\
\hline
\multirow{3}{4em}{\textbf{Ours}} & ResNet-18 & 92.3 & 94.6 \\
 & ViT-B/32 & \textbf{93.4}& \textbf{96.3} \\
 & \cellcolor{green!25}ViT-B/16 & \cellcolor{green!25}\textbf{94.2}& \cellcolor{green!25}\textbf{96.7} \\
\hline
\end{tabular}

\end{center}

\label{table:SOTA_Volleyball}

\end{table}

\begin{table}[!t]

\makeatletter\def\@captype{table}

\centering
\setlength{\tabcolsep}{0.8mm}
\captionof{table}{Comparative results of the group activity recognition on JRDB-PAR dataset~\cite{han2022panoramic}. $\mathcal{P}_g$, $\mathcal{R}_g$, and $\mathcal{F}_g$ represent Precision, Recall and F1-score. In these experiments, we use ViT-B/16 as the backbone. }
{\begin{tabular}[*c]{l|c|c|c}
				\hline
				\multirow{2}{30pt}{\textbf{Method}}  &\multicolumn{3}{c}{\textbf{Classification}}  \\   \cline{2-4} 
				&\textbf{$\mathcal{P}_g$}   &\textbf{$\mathcal{R}_g$}  &\textbf{$\mathcal{F}_g$} 
				 \\ \hline 
                \multicolumn{4}{c}{\textbf{Fully supervised}} \\
                \hline
                PCTDM~\cite{yan2018participation}                & 32.1   & 27.2  & 28.9\\
				ARG~\cite{wu2019learning}  &34.6 &29.3 &30.7 \\
                HiGCIN~\cite{yan2020higcin}                      & 35.1            & 34.4  & 31.8 \\
                AT\cite{gavrilyuk2020actor} 	&34.8 	&37.6 &	 30.9\\
				SA-GAT~\cite{ehsanpour2020joint} 	&36.7 &29.9 &31.4 \\		
				JRDB-Base~\cite{ehsanpour2022jrdb} 	&44.6	&46.8	&45.1	\\		
                SoGAR~\cite{chappa2023sogar} &49.3&47.1&48.7\\
				\textbf{Ours}   	&\textbf{50.1} 	&\textbf{48.9} 	&\textbf{49.2}	 \\
				\hline   
                \multicolumn{4}{c}{\textbf{Weakly supervised}} \\
                \hline 			
				AT\cite{gavrilyuk2020actor} 	&21.2 	&19.1 &19.8	 \\
				SACRF\cite{pramono2020empowering} 	&42.9 	&35.5  &37.6 \\
				Dynamic\cite{yuan2021spatio} 	&37.5 &27.1 &30.6	\\
				HiGCIN\cite{yan2020higcin} 	&39.3 	&30.1 &33.1	  \\		
				ARG\cite{wu2019learning} 			&26.9 	&21.5 & 23.3 \\
				SA-GAT\cite{ehsanpour2020joint}  	&28.6 	&24.0 	&25.5 \\	
				JRDB-Base\cite{ehsanpour2022jrdb} 	&38.4	&33.1	&34.8\\
				SoGAR~\cite{chappa2023sogar} &47.1&45.8&44.9\\
                \textbf{Ours}  	&\textbf{48.3} 	&\textbf{47.5} 	&\textbf{47.2} \\
				\hline      
			\end{tabular}}
   
\label{tab:jrdbpar}


\end{table}
\noindent\textbf{Volleyball dataset.} We assess the effectiveness of our approach in comparison to state-of-the-art methods in both Group Activity Recognition (GAR) and Weakly Supervised Group Activity Recognition (WSGAR) across two supervision levels: fully supervised and weakly supervised. The former involves using actor-level labels, including ground-truth bounding boxes and individual action class labels during both training and inference. For equitable comparisons, we provide results of previous methods and reproduce outcomes using RGB input and a ResNet-18 backbone.
In the weakly supervised setting, we replace group action classification labels with ground-truth bounding boxes of actors, excluding their corresponding actions. This allows the model to learn actor localization during the pre-training stage. Table~\ref{table:SOTA_Volleyball} summarizes the results, with the first section depicting outcomes of prior techniques in fully supervised conditions and the second section in weakly supervised scenarios.
Our model, trained on the ResNet-18 backbone, surpasses most fully supervised frameworks, notably enhancing MCA and MPCA metrics. With the ViT-Base backbone, our approach excels in weakly supervised conditions, outperforming all GAR and WSGAR models. Leveraging spatiotemporal features through the transformer architecture, it achieves a substantial improvement of 2.4\% in MCA and 1.2\% in Merged MCA. Furthermore, our approach outperforms current GAR methods that rely on less comprehensive actor-level supervision, as observed in~\cite{bagautdinov2017social, yan2018participation, qi2018stagnet, gavrilyuk2020actor, pramono2020empowering}.

\subsection{Human Action Retrieval Evaluation}

Our experimental results in Table.~\ref{tab:objectretrieval} show that our proposed method outperforms the existing state-of-the-art approaches in terms of the R@K metric. Specifically, our method achieved a significantly higher R@1 score than the other methods, with a margin of \textit{4.8}. Additionally, our method achieved higher R@5 and R@10 scores than all other methods, demonstrating its superior performance in video object retrieval. Furthermore, our method demonstrated a more consistent performance across two datasets and experimental setups. We attribute this to our method's ability to effectively capture the video frames' global and local features and its incorporation of text embeddings for improved semantic understanding of the video content.

\begin{table}[!t]
\small
\setlength{\belowcaptionskip}{1pt}
\centering
\caption{Human Action-retrieval task results on transformer-based methods.}
\setlength{\tabcolsep}{0.45mm}{
\begin{tabular}{cccccccc}
\toprule
\multirow{3}{*}[5pt]{\textbf{Method}}  & \multirow{3}{*}[5pt]{\textbf{Backbone}} &  \multicolumn{3}{c}{\textbf{Volleyball}}  & \multicolumn{3}{c}{\textbf{JRDB-PAR}}   \\  
\cmidrule(r){3-5}\cmidrule(r){6-8}
& & R@1& R@5 & R@10 & R@1 & R@5 &  R@10 \\ 
\midrule
AT\cite{gavrilyuk2020actor} & \multirow{4}{*}{ResNet18} & 61.2 & 79.2 & 95.2 & 46.3 & 54.4 & 74.4  \\
DFWSGAR~\cite{kim2022detector} & & 64.7 & 80.4 & 85.3 & 47.6 & 57.9 & 77.1 \\
SPARTAN~\cite{chappa2023group} & & 68.6 & 80.6 & 95.7 & 54.6 & 60.1 & 80.2 \\ 
\textbf{Ours}& & \textbf{71.3} & \textbf{84.5} & \textbf{96.3} & \textbf{61.4} & \textbf{71.3} & \textbf{87.4} \\ 
\midrule
AT\cite{gavrilyuk2020actor} & \multirow{4}{*}{ViT-B/32}& 54.2 & 77.6 & 93.6 & 41.5 & 49.2 & 68.7 \\
DFWSGAR~\cite{kim2022detector} & & 57.6 & 59.8 & 84.4 & 44.2 & 53.6 & 70.8  \\
SPARTAN~\cite{chappa2023group} &  & 58.1 & 81.3 & 95.4 & 49.5 & 59.2 & 74.3 \\ 
\textbf{Ours}& & \textbf{64.4} & \textbf{84.2} & \textbf{95.4} & \textbf{55.6} & \textbf{67.2} & \textbf{81.3} \\
\midrule
AT\cite{gavrilyuk2020actor} & \multirow{3}{*}{ViT-B/16}& 62.3 & 84.6 & 97.3 & 50.2 & 58.1 & 77.2  \\
DFWSGAR~\cite{kim2022detector} & & 65.7 & 59.8 & 84.4 & 54.3 & 62.7 & 79.8  \\
SPARTAN~\cite{chappa2023group} &  & 69.4 & 84.5 & 97.4 & 59.1 & 65.3 & 82.1 \\ 
\cellcolor{green!25}\textbf{Ours}& \cellcolor{green!25}& \cellcolor{green!25}\textbf{74.2} & \cellcolor{green!25}\textbf{85.6} & \cellcolor{green!25}\textbf{97.7} & \cellcolor{green!25}\textbf{67.2} & \cellcolor{green!25}\textbf{71.4} & \cellcolor{green!25}\textbf{88.2}  \\
\bottomrule

\end{tabular}}

\label{tab:objectretrieval}
\end{table}

\subsection{Ablation Study}
In this section, we verify the effectiveness of each component in the REACT framework on downstream group activity recognition tasks and verify the significance of the actor fusion block on the proposed method. In Table.~\ref{tab:garablation}, irrespective of the visual backbone, results indicate that all the components, including feature extraction, encoding, and decoding, can bring significant gains individually. From Table.~\ref{tab:ablation}, results show the importance of actor fusion block in all the experiments. More ablation results can be seen in {Appendix}, including the detailed ablation on the effectiveness of each component on downstream human action retrieval tasks. 

 \begin{table}[!t]

\vspace{-1pt}
\captionof{table}{Ablation study of each Group Activity Recognition task component. \textbf{VB} denotes Visual Backbone and \textbf{FE} denotes Feature Extraction.}
\makeatletter\def\@captype{table}
\small
\centering
\setlength{\tabcolsep}{0.5mm}{\begin{tabular}[*c]{c|c|c|c|c|c}
\toprule
\small\textbf{\small VB} & \textit{\textbf{\small FE}} & \textit{\textbf{\small Encoding}} &\textbf{\small Decoding} & \textbf{\small JRDB-PAR} & \textbf{\small Volleyball}  \\ \midrule
\multirow{4}{*}{ResNet-18} & \cmark & & & 36.5 & 82.4  \\
& \cmark& \cmark& & 41.3(\textcolor{darkergreen}{+4.8}) & 86.5(\textcolor{darkergreen}{+4.1}) \\ 
& \cmark&\cmark &\cmark & \textbf{46.4}(\textcolor{darkergreen}{+5.1}) & \textbf{92.3}(\textcolor{darkergreen}{+5.8}) \\ 
\midrule
\multirow{4}{*}{ViT-B/32} & \cmark & &  & 31.8 & 80.2\\
& \cmark&\cmark & & 39.2(\textcolor{darkergreen}{+7.4}) & 84.3(\textcolor{darkergreen}{+4.1})\\
&\cmark&\cmark & \cmark& \textbf{45.7}(\textcolor{darkergreen}{+6.5}) &\textbf{93.4}(\textcolor{darkergreen}{+9.1})\\

\bottomrule
\multirow{4}{*}{ViT-B/16} & \cmark & &  & 38.3 & 79.7\\
& \cmark&\cmark & & 41.4(\textcolor{darkergreen}{+3.1}) & 85.7(\textcolor{darkergreen}{+6.0})\\
&\cmark&\cmark & \cmark& \textbf{47.2}(\textcolor{darkergreen}{+6.2}) &\textbf{96.7}(\textcolor{darkergreen}{+11.0})\\

\bottomrule
\end{tabular}}

\label{tab:garablation}
\end{table}
\hfill
\begin{table}
    \centering

\caption{Ablation study results on the Actor Fusion component in the architecture. For all the experiments below, we use ViT-B/32 as the visual backbone for the framework. The metrics we use for the Volleyball dataset are \textit{Merged MCA}, and for the JRDB-PAR dataset \textit{$\mathcal{F}_g$}. }\label{tab:ablation}

\begin{tabular}{c|c|c}
\hline
   \textbf{Actor Fusion} & \textbf{JRDB-PAR} & \textbf{Volleyball} \\
\hline
  \xmark & 33.7 & 76.1\\  
 \cellcolor{green!25}\cmark & \cellcolor{green!25}\textbf{47.2} & \cellcolor{green!25}\textbf{96.7}\\
\hline
\end{tabular}

\end{table}
\subsection{Visualization}

\begin{figure}[!ht]
    \centering
    \includegraphics[width=0.48\textwidth]{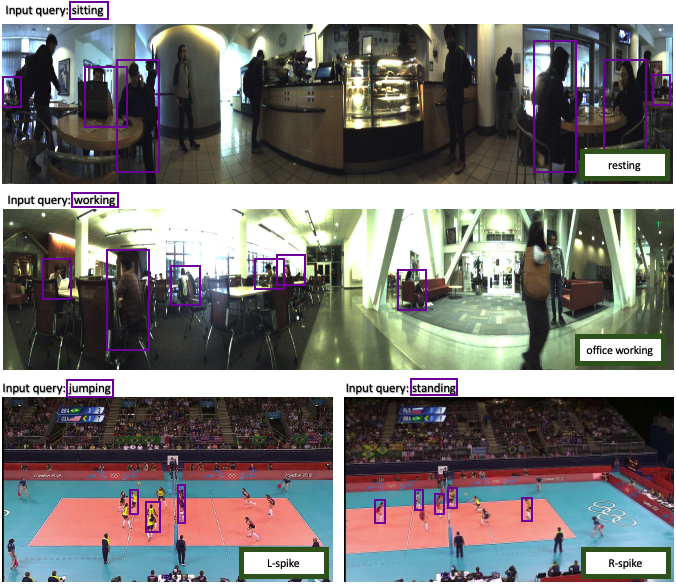}
    \caption{
    {\textbf{Visualization based on input action query.}The top two rows are the results from the JRDB-PAR dataset, and the bottom row is from the Volleyball dataset. \textbf{Best viewed in color and zoom in.}} }
    \label{fig:vis1}
\end{figure}

In order to gain insights into the visualization capabilities of our proposed method and demonstrate its significance, we conducted an analysis that involved examining the visualizations based on the input action query. The visualizations are presented in Fig.~\ref{fig:vis1}, which showcases the localization of actions based on the input query for the JRDB-PAR and Volleyball datasets. The results of the decoder's generated proposals during the inference stage indicate that our method effectively localizes the specified actions. It demonstrates the ability of our framework to accurately recognize and pinpoint specific actions within a group activity context.
Furthermore, the t-SNE plots are visualized in the {Appendix}. The results illustrated in these plots demonstrate that the proposed framework effectively recognizes all actions within a given video. Moreover, the framework enhances the overall performance in group activity recognition.

\section{Conclusion}

 We introduce REACT, a novel video transformer-based model for group activity recognition using contrastive learning. Our approach generates diverse spatio-temporal views from a single video, leveraging different scales and frame rates. Correspondence learning tasks capture motion properties and cross-view relationships between sampled clips and textual information.
The core of our approach is a contrastive learning objective that reconstructs video-text modalities in the latent space. REACT effectively models long-range spatio-temporal dependencies and performs dynamic inference within a single architecture.
REACT outperforms state-of-the-art models in group activity recognition, showcasing its superior performance.

\noindent\textbf{Limitations.} 
While our methods excel in group activity classification and human action-retrieval tasks, they may need to be optimized for other visual-language problems due to the lack of textual descriptions in the datasets. Future research directions could focus on incorporating additional data involving question-answering, phrases for video grounding, and other related textual information to enhance the versatility of our methods.
{
    \small
    \bibliographystyle{ieeenat_fullname}
    \bibliography{main}
}



\end{document}